\documentclass[11pt,a4paper]{article}
\usepackage{times,latexsym}
\usepackage{url}
\usepackage[T1]{fontenc}
\usepackage[most]{tcolorbox}
\usepackage[acceptedWithA]{tacl2021v1}
\usepackage{graphicx}
\usepackage{amsmath}
\usepackage{amssymb}
\usepackage{booktabs}
\usepackage{algorithmic}
\usepackage{amsfonts}
\usepackage{subfigure}
\usepackage{url}
\usepackage{multirow}
\usepackage{multicol}
\usepackage{arydshln}
\usepackage{bbm}
\usepackage{bm}        

\usepackage[ruled,vlined]{algorithm2e}
\usepackage{colortbl}
\usepackage{tabularx}
\definecolor{celestialblue}{rgb}{0.29, 0.59, 0.82}
\definecolor{amber}{rgb}{1.0, 0.75, 0.0}
\definecolor{palechestnut}{rgb}{0.87, 0.68, 0.69}
\usepackage{listings}

\newcommand{\agentname}
{PLAYER*}

\definecolor{yellow}{HTML}{F6BD60}
\definecolor{white}{HTML}{F7EDE2}
\definecolor{pink}{HTML}{F5CAC3}
\definecolor{tale}{HTML}{84A59D}
\definecolor{red}{HTML}{F28482}
\definecolor{green}{HTML}{8EC07C}
\definecolor{orange}{HTML}{FE8019}
\definecolor{grey}{HTML}{EBDBB2}
\definecolor{brain}{HTML}{FFABBE}
\definecolor{blue}{HTML}{076678}
\definecolor{narrative}{HTML}{458588}

\usepackage{xspace,mfirstuc,tabulary}

\newif\iftaclinstructions
\taclinstructionsfalse 
\iftaclinstructions
\renewcommand{\confidential}{}
\renewcommand{\anonsubtext}{(No author info supplied here, for consistency with
TACL-submission anonymization requirements)}
\newcommand{\instr}
\fi

\iftaclpubformat 

\else

\fi

\title{PLAYER*: Enhancing LLM-based Multi-Agent Communication and Interaction in Murder Mystery Games}

\author{Qinglin Zhu$^{1*}$, Runcong Zhao$^{1*}$, Bin Liang$^{4,5}$, Jinhua Du$^2$, \textbf{ Lin Gui$^1$}, Yulan He$^{1,3}$\\
  $^1$King's College London, 
  $^2$Huawei London Research Centre, 
  $^3$The Alan Turing Institute,\\
  $^4$The Chinese University of Hong Kong, 
  $^5$MoE Lab, CUHK\\
  \texttt{\{qinglin.1.zhu,runcong.zhao\}@kcl.ac.uk}, 
  \texttt{bin.liang@cuhk.edu.hk}\\
  \texttt{\{jinhua.du\}@huawei.com},  
  \texttt{\{lin.1.gui,yulan.he\}@kcl.ac.uk} }

\begin{document}
\maketitle

\begin{abstract}
We introduce WellPlay, a reasoning dataset for multi-agent conversational inference in Murder Mystery Games (MMGs). WellPlay comprises 1,482 inferential questions across 12 games, spanning objectives, reasoning, and relationship understanding, and establishes a systematic benchmark for evaluating agent reasoning abilities in complex social settings. Building on this foundation, we present \agentname, a novel framework for Large Language Model (LLM)-based agents in MMGs. MMGs pose unique challenges, including undefined state spaces, absent intermediate rewards, and the need for strategic reasoning through natural language. \agentname\ addresses these challenges with a sensor-based state representation and an information-driven strategy that optimises questioning and suspect pruning. Experiments show that \agentname outperforms existing methods in reasoning accuracy, efficiency, and agent-human interaction, advancing reasoning agents for complex social scenarios.
\end{abstract}

%

\section{Introduction}
Recent advancements in LLMs capable of generating human-like responses have boosted the development of LLM-based agents \citep{llm-2023-soni, furchat-2023-cherakara}. Building on this progress, a series of studies focusing on multi-agent communications have showcased the emergence of social interactions, including cooperation \citep{camel-li2023, diplomacy-2022fair}, trust \citep{werewolf-xu2023}, deception \citep{avalon-wang2023}, and information propagation \citep{generative-agent-park2023}. 
However, the study of multi-agent \textbf{complex social reasoning}—where agents must infer others’ beliefs, plans, and hidden goals in dynamic environments—remains significantly underexplored. One major bottleneck is the lack of large-scale, high-quality benchmarks capturing such intricate interactions. Furthermore, current evaluation methods either focus solely on the final win/loss outcome or rely heavily on manual evaluation, which significantly limits the scope of analysis due to the high cost 
\citep{werewolf-xu2023,  murdergame-wu2024}. 

\begin{figure}[t]
    \centering
    \includegraphics[width=\linewidth]{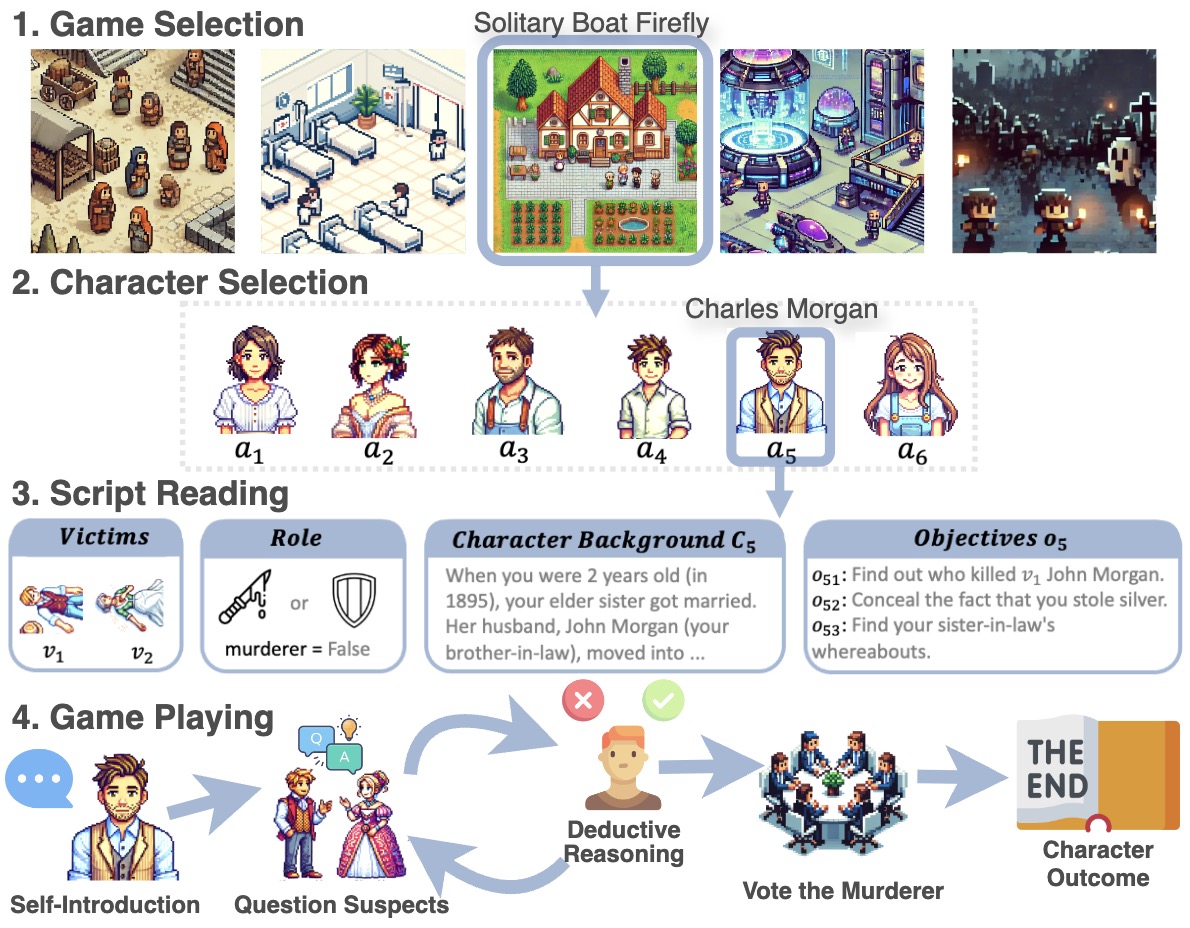}
    \caption{The murder mystery game involves players taking on roles, questioning suspects, and using deduction to identify the killer, all while pursuing their character-specific objectives.
    }
    \label{fig:overview}
\end{figure}

To bridge this gap, we adopt the murder mystery game (MMG) as a testbed to investigate whether LLMs are able to simulate complex social reasoning. 
As shown in Figure~\ref{fig:overview}, MMGs are strategic games involving 4–12 players assuming character roles with specific scripts and objectives. Gameplay unfolds through language-based negotiation and tactical coordination over $n$ rounds, with players limited to $m$ questions per round. The ultimate goal is to uncover the truth collaboratively while advancing individual objectives. At the end of the game, players vote on the suspected killer and evaluate objective completion.


We choose MMG as the use case for three reasons:  
1) \textbf{Social Behaviour-based Evaluation}: MMGs are script-based character role-playing games that contain complex social relationships across different topics. Compared to current reasoning tasks, which focus on open question answering or mathematical problem solving, MMGs cover a variety of topics. They are more suitable for evaluating whether the LLM understands complex human social behaviours.  
2) \textbf{Social Strategy Understanding}: MMGs involve multiple goals for a single character. Winning or losing is never the sole objective for role-playing, as a character might assist his or her relative(s) who is the murderer in concealing their crime. These varying agent goals are dependent on the script and require the LLM to understand many important concepts in human social strategy, such as trust, cooperation, deception, and selfishness.  
3) \textbf{Social Activities Involvement}: MMGs enable human evaluation by allowing participants to engage directly in the game, rather than simply assessing the quality of generation and reasoning from LLMs.


To investigate the \textbf{Social Behaviour-based Evaluation}, we introduce \textbf{WellPlay}, a new dataset containing 1,482 inferential questions across 12 diverse Murder Mystery Games (MMGs). WellPlay categorises questions into three types: objective, reasoning and relation, enabling a more comprehensive and scalable evaluation of agents' reasoning capabilities beyond binary outcomes.

To investigate the \textbf{Social Strategy Understanding}, we propose \textbf{PLAYER*}, an LLM-driven heuristic search framework in which the LLM dynamically decides whom to ask and what to ask in a multi-agent environment. Each character is represented by a sensor-based state capturing mental and situational attributes, updated through natural language prompting. Agents explore this state space via generative question-answering, framed as a planning problem. To guide the search, we introduce an Information Gain (IG)-based heuristic that balances current search results with the potential value of future queries. After each interaction, agents assess whether the gathered information updates character states and narrows down suspects, enabling a targeted selection of the next questioning action based on IG.



To investigate the \textbf{Social Activities Involving}, we recruited human players to interact directly with the agent. The results revealed that previous agents often overfit to agent-vs-agent interaction environments, exhibiting verbose and repetitive dialogues that detracted from the overall player experience. By integrating our goal-oriented pruner, these issues were significantly mitigated, resulting in more natural and human-like agent behaviour and greatly enhancing player satisfaction.

In summary, our contributions are as follows: 1) the construction of a dataset with an evaluation framework; 2) a flexible yet efficient framework to assess the complex social reasoning abilities of LLMs; 3) a human-agent interaction platform.\footnote{Our code and dataset are available at \url{https://github.com/alickzhu/PLAYER}, along with detailed MMG rules and procedures.}

\section{The WellPlay Dataset} 


\begin{table*}[ht!]
\centering
\resizebox{0.85 \linewidth}{!}{
\begin{tabular}{ccccrrrcccc}
\toprule
\multicolumn{1}{c}{\multirow{2}{*}{\textbf{MMG}}} & \multicolumn{1}{c}{\multirow{2}{*}{\textbf{\#Agents}}} & \multicolumn{1}{c}{\multirow{2}{*}{\textbf{\#Victims}}} & \multicolumn{2}{c}{\textbf{\#token(CN)}} & \multicolumn{2}{c}{\textbf{\#token(EN)}} &  \multicolumn{4}{c}{\textbf{Question}}    \\ 
\cmidrule(lr){4-5}  \cmidrule(lr){6-7}\cmidrule(lr){8-11}
&         &              & \textbf{avg}& \textbf{overall}&\textbf{ avg}& \textbf{overall}& \textbf{Objective}& \textbf{Reasoning}& \textbf{Relations}& \textbf{overall}\\

\midrule
\emph{Death Wears White}        & 9 & 1 & 3,191 & 28,716  & 1,742 & 15,681 & 10 & 102 & 72  & 184 \\
\emph{Ghost Revenge}            & 7 & 3 & 5,488 & 38,415  & 3,960 & 27,723 & 19 & 152 & 69  & 240 \\
\emph{Danshui Villa}            & 7 & 2 & 5,111 & 35,779  & 3,339 & 23,370 & 12 & 128 & 63  & 203 \\
\emph{Unfinished Love}          & 7 & 2 & 2,501 & 17,507  & 1,652 & 11,562 & 12 & 61  & 72  & 145 \\
\emph{Cruise Incident}          & 5 & 1 & 1,263 & 6,313   & 808  & 4,040  & 4  & 24  & 30  & 58  \\
\emph{Sin}                      & 4 & 1 & 2,121 & 8,485   & 1,378 & 5,512  & 3  & 20  & 21  & 44  \\
\emph{Deadly Fountain}          & 4 & 1 & 1,852 & 7,410   & 1,194 & 4,775  & 3  & 21  & 12  & 36  \\
\emph{Unbelievable Incident}    & 5 & 1 & 3,182 & 15,912  & 2,012 & 10,062 & 4  & 24  & 15  & 43  \\
\emph{Desperate Sunshine}       & 4 & 1 & 3,370 & 13,481  & 2,219 & 8,874  & 3  & 18  & 36  & 57  \\
\emph{Riverside Inn}            & 4 & 1 & 1,910 & 7,638   & 1,257 & 5,028  & 3  & 18  & 18  & 39  \\
\emph{Solitary Boat Firefly}    & 6 & 4 & 8,894 & 53,362  & 6,874 & 41,244 & 20 & 109 & 69  & 198 \\
\emph{Manna}                    & 6 & 3 & 9,028 & 54,169  & 6,492 & 38,954 & 24 & 123 & 88  & 235 \\
\midrule
\textbf{Avg}                              & 5.67 & 1.75 & 3,993 & 23,932 & 2,744 & 16,402 & 9.75 & 66.67 & 47.08 & 125.50 \\
\textbf{Sum}                              & 68 & 21 & 47,911 & 287,187 & 32,927 & 196,825 & 117 & 800 & 565 & 1482 \\

\bottomrule
\end{tabular}
}
\caption{Dataset Statistics. \textit{Agents} is the count of players, \textit{Victims} is the number of victims, \#\textit{token(CN)} and \#\textit{token(EN)} are the token counts in the Chinese and English dataset versions, respectively. \textit{Avg} shows the average script length per character, \textit{Overall} is the total script token count, and \textit{Question} enumerates the number of questions by types. The number of evaluation questions varies based on script complexity, with more complex scripts generating a larger volume of questions. }
\label{tab:dataset_statistic}
\end{table*}

\paragraph{Dataset Setting}
WellPlay is a dataset constructed from MMG scenarios, designed to evaluate multi-agent complex social reasoning abilities.
Each scenario defines a set of agents \(\mathcal{A} = \{a_i\}_{i=1}^{N_a}\) and a set of victims \(\mathcal{V} = \{v_k\}_{k=1}^{N_v}\), where \( N_a \) and \( N_v \) denote the numbers of playable characters and victims, respectively. Each agent \(a_i\) is initialised with the following: \\
(1) \textbf{Role $\mathbf{r}_i = \{r_{ik}\}_{k=1}^{N_v}$}: Whether or not they are the murderer of each victim;\\
(2) \textbf{Role background script \(C_i\)}: Crafted from the unique viewpoint of \(a_i\); \\
(3) \textbf{Objectives 
$\mathbf{o}_i = \{o_{ij}\}_{j=1}^{N_{o_{i}}}$}: 
A set of \( N_{o_i} \) goals assigned to agent \( a_i \) within the game.

In addition, Game Rules are also provided to Agents as essential information. 

\paragraph{Dataset}
While the original MMG scripts provide basic annotations of character relationships \citep{conan-zhao2024}, they are insufficient for evaluating an agent’s understanding of the full game context, situational reasoning, and decision-making abilities.
To address this limitation, we designed a comprehensive set of evaluation questions and annotated the dataset accordingly. 
We employ multiple-choice questions focusing on \emph{factual information} to ensure a quantifiable evaluation and minimise controversy. We have recruited four annotators to label inferential questions on:

(1) \emph{Objective}. Including shared objectives, such as identifying the perpetrator(s), and individual objectives, such as determining who stoles the wallet, for each character in the game.

(2) \emph{Reasoning}. This entails questions that delve into the reasoning behind provided answers, relating to agents' objectives, including:
Who; 
What (the nature of the incident, such as murder, theft, or disappearance);
When (the time of the incident);
Where(the location of the incident);
Cause (e.g., shooting, poisoning, stabbing); 
Motive (e.g., crime of passion, vendetta, or manslaughter).

(3) \emph{Relations}. This includes interpersonal relationships between victims and others, as well as relationships among suspects, with labels adapted from the Conan dataset \citep{conan-zhao2024}.

WellPlay encompasses 12 MMGs, comprising a total of 1,482 evaluation questions (examples presented in Table~\ref{tab:question_example}). On average, each game features 5.67 agents and 1.75 victims (see Table~\ref{tab:dataset_statistic}).  

\section{\agentname \ }


\begin{figure*}[h!]
    \centering
    \includegraphics[width=\linewidth]{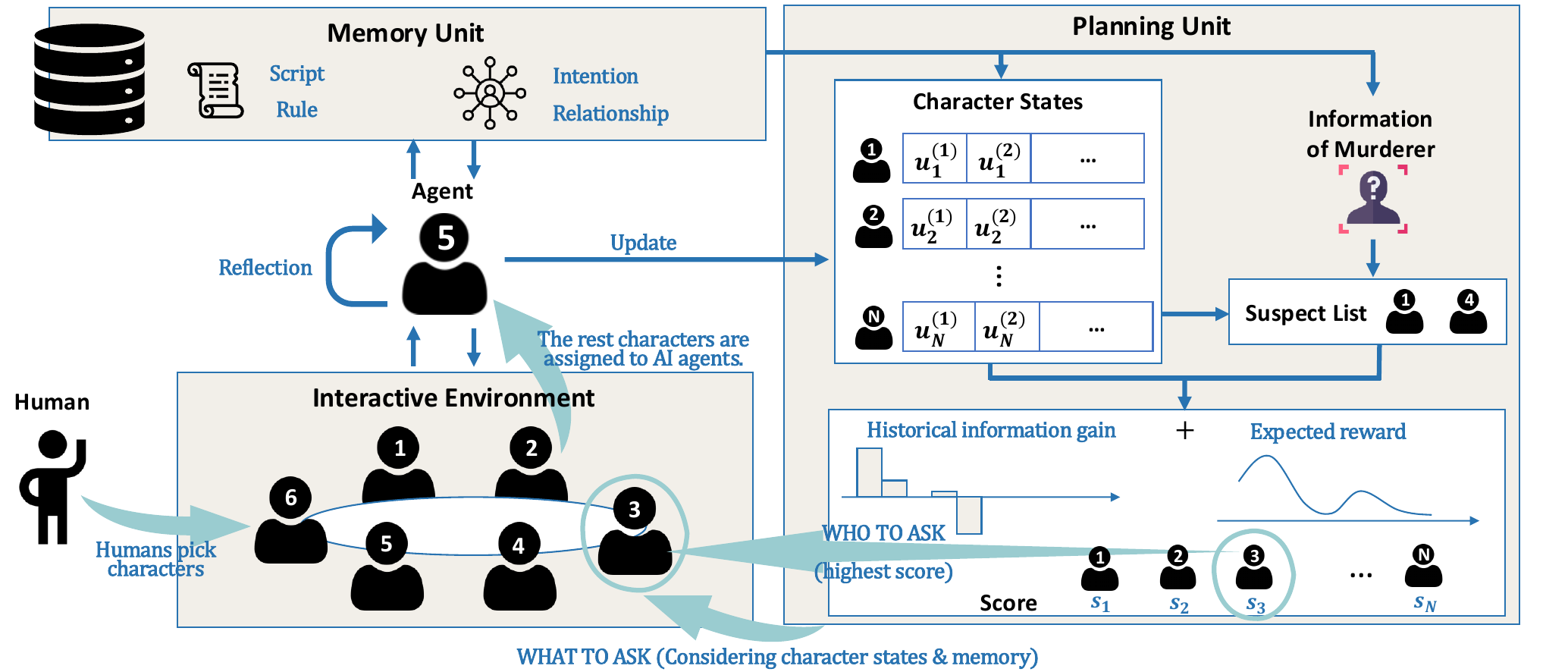}
    \caption{\textbf{Search and Approximate.} 
    \agentname \  generates questions based on character states, selecting agents to question based on past observations of critical information and the likelihood of uncovering more. The goal is to minimise the suspect list.}
    \label{fig:search-approximate}
\end{figure*}

\agentname \ approximates the search domain through sampling, and plans the shortest path to the agent's objective by prioritising searches based on the quality of potential solutions. As illustrated in Figure~\ref{fig:search-approximate}, this framework is fundamentally composed of two key components: \emph{(1) Search via sensor-based state matching:} \agentname \  searches for the murderer based on their proximity to the ideal murderer in the language space.
\emph{(2) Approximation with a pruner:} \agentname \  focuses on a subset of characters that are highly suspicious of being a murderer, and decides whom to query based on this suspect list.


In MMGs, the suspect status is \( \mathbf{s_i} = \left[ s_{ijk} \right]_{j=1, k=1, j \neq i}^{N_a, N_v} \), where $s_{ijk} \in [0, 1]$. Here, \( s_{ijk} \) encodes agent \( a_i \)'s beliefs about player $a_j$'s likelihood of being victim \( v_k \)'s murderer. In our setting, $s_{ijk}$ is set to be discrete ($\in \{0,1\}$), representing suspect and non-suspect.
While our list of suspects is finite, our actual search space $L \in \mathbb{R}^d$ is continuous and high-dimensional, which encodes character-level semantics—such as emotion, motivation, and relationships—derived from language model embeddings.
It encompasses the dynamics of relationships, perceptions towards the agent, and hidden secrets.

\paragraph{Search via Sensor-based State Matching}

To effectively model suspects within a continuous language domain, we introduce a set of \textbf{domain-specific sensors} designed to capture essential attributes of each agent (player). Specifically, we represent each agent \( a_i \) using a vector \( \mathbf{u}_i = (e_i, m_i, p_i, \dots) \) in the state space \( L \), where each dimension reflects a distinct sensor reading. These sensors include \emph{emotion} (\( e_i \)), which assesses an agent's feelings about the crime or other characters; \emph{motivation} (\( m_i \)), indicating potential incentives such as revenge or greed; and \emph{opportunity} (\( p_i \)), determining whether the agent could realistically commit the crime. The selection of these attributes leverages established sociological and psychological theories \citep{social-impact-1981latane, self-enhancement-bias-2021zhao, psychological-distance-2010trope}, ensuring they capture attributes crucial for crime-solving.

Initially, we establish a murderer profile \( \mathbf{h}_k \) for each victim \( v_k \), using narrative priors reflecting attributes like strong motive, negative emotion, and sufficient opportunity. This profile dynamically evolves as the game progresses and new evidence emerges.
We then define the binary suspect status of agent \( a_j \) regarding victim \( v_k \) as follows: $s_{jk} = \text{LLM}(\mathbf{u}_j, \mathbf{h}_k)$,
indicating whether the agent's sensor-based state aligns sufficiently with the inferred murderer profile.

Our method involves iteratively \textbf{sampling} and \textbf{updating} each suspect's state to fulfill two core objectives: (1) \textbf{Profile Alignment:} Estimating how closely each suspect’s attributes align with the evolving profile of the murderer. After each round of questioning or discovery, sensor readings are updated to reflect new evidence, recalculating each agent’s alignment with the murderer profile.
(2) \textbf{Targeted Inquiry:} Generating precise questions to resolve ambiguities in suspect states. We formulate the question generation process as \( q = g(\mathbf{u}_i, X) \), where \( g \) is a language model informed by relevant context \( X \), obtained through Retrieval-Augmented Generation (RAG). By concentrating on uncertain or suspicious sensor values—such as unclear motives or contradictory emotional signals—the agent strategically designs questions that reduce state ambiguity and efficiently narrow down potential suspects.

\paragraph{Approximation with a pruner}
Following the questioning step, agents receive responses from the questioned agent as well as dialogue among other agents. Each agent then updates its beliefs about the murderer, as well as the character states of other agents. In human gameplay, players often form judgments regarding the scenario and focus on a subset of highly suspicious individuals. This strategic refinement process is crucial for conserving cognitive resources and enhancing the efficacy of inferential reasoning. Inspired by the human thinking process, we also enable agents to consider the suspect list. 

In an MMG with $N_a$ players, to find out who is the murderer $Y$ of a given victim $v_k$, we define the entropy to be 
$H = -ln(\frac{1}{n})$, where $n$ is the number of suspects in the agent's suspect list. Based on the size of the suspect list for each round of the game, the agent can calculate the corresponding entropy. The target is to minimise this entropy. 
The initial entropy is $H_0 = -ln(\frac{1}{N_a-1})$, where everyone except the agent itself, who aims to find the murderers, are all suspects, representing the uncertainty about who the murderer is at the start of the game. 
At each round \( i \), we select only one character for questioning. The information gain for the selected character \( c^* \) in this round is: 
\begin{equation}
IG_{i, c^*} = H_{i-1} - H_i.
\end{equation}
This value quantifies the reduction in uncertainty about the murderer's identity after questioning \( c^* \). If questioning this character reduces the suspect list, then \( H_i < H_{i-1} \), making \( IG_{i, c^*} \) positive, indicating useful information was obtained. Conversely, if questioning does not lead to any reduction in the suspect list, then \( IG_{i, c^*} \) is zero or negative, implying no valuable information was gained, and potentially even introducing noise into the reasoning process.

For all other characters who were not questioned in this round (\( c \neq c^* \)), the information gain remains zero:
$IG_{i, c} = 0, \forall c \neq c^*.$

The accumulated historical IG for character $c$ across rounds is: $IG_{1:i, c} = [IG_{1, c}, ..., IG_{i, c}].$ 

The weighted historical IG is:
\begin{equation}
IG_{\text{weighted}, c}[i] = \frac{\sum_{j < i} w_j \cdot IG_{j, c}}{\sum_{j < i} w_j},
\end{equation}
where the weight \( w_j = e^{-(i - j)} \) ensures that more recent rounds contribute more significantly to the final score.
Open-source projects such as Llama could introduce strong bias as they usually focus on one suspect \cite{DBLP:journals/corr/abs-2310-18940}. To mitigate this, we propose a heuristic estimator based on prompting. Specifically, we first prompt the LLM to assess whether questioning a given character would yield information gain. By directly leveraging the LLM's probability distribution, we obtain an expected information gain in the range \([0,1]\):
\begin{equation}
E[IG_{i+1}] =
\begin{cases}
p_{\text{yes}}, & \text{if LLM returns ``yes''} \\
1 - p_{\text{no}}, & \text{if LLM returns ``no''}
\end{cases}
\end{equation} 
where \( p_{\text{yes}} \) and \( p_{\text{no}} \) represent the probabilities assigned by the LLM to the ``yes'' and ``no'' responses, respectively.
We then integrate historical and expected information gain to determine who to ask in the next round. The score for each candidate character \( c \) is:
\begin{equation}
\text{Score}_c = \beta IG_{\text{weighted}}[i] + (1-\beta) E[IG_{i+1, c]},
\end{equation}
The agent selects the character with the highest score:
\begin{equation}
c^* = \arg\max_{c} \text{Score}_c.
\end{equation}
In addition, we also implement an $\epsilon$-greedy strategy to choose a random character with probability $\epsilon$, and the highest-scoring character with probability $1-\epsilon$ to balance exploration and exploitation. 
The detailed procedure is in Algorithm~\ref{alg:player}.

\begin{algorithm}[th]
\algsetup{linenosize=\tiny}
\small
\caption{\agentname\ Framework} \label{alg:player}
\LinesNumbered
\KwIn{Agents $\mathcal{A} = \{a_i\}_{i=1}^{N_a}$, Victims $\mathcal{V} = \{v_k\}_{k=1}^{N_v}$, Max round $n$}
\KwOut{Evaluation of Results}

$current\_round \gets 0$ \\
\textbf{Init:} Character states $\mathbf{u}$, Suspicious states $\mathbf{s}$ \\

\While{$current\_round \leq n$}{
    \tcp{Search via state matching}
    \For{$i = 1$ \KwTo $N_a$}{
        \For{$k = 1$ \KwTo $N_v$}{
            $suspect\_list_{ik} \gets$ \texttt{Suspect\_Generation}($s_{ik}$) \\

            \For{each $a_j \in \mathcal{A} $}{
                $score_j \gets \beta \cdot \texttt{HistoricalIG}(a_j) + (1 - \beta) \cdot \texttt{ExpectedIG}(a_j)$
            }
            $a_j^* \gets \arg\max_{a_j} score_j$ \\

            $q \gets$ \texttt{GenerateQuestion}($a_j^*, v_k, \mathbf{u}_{j^*}$) \\
            $r \gets$ \texttt{GetReply}($a_j^*, q$) \\
            \texttt{UpdateMemory}($a_i, r$)
        }
    }

    \tcp{Approximation with a pruner}
    \For{$i = 1$ \KwTo $N_a$}{
        \For{$k = 1$ \KwTo $N_v$}{
            $\mathcal{M}_{ik} \gets$ \texttt{RetrieveMemory}($a_i, v_k$) 
            $s_{ik} \gets$ \texttt{Update\_s}($a_i, v_k, \mathcal{M}_{ik}$)
        }
    }

    $current\_round \gets current\_round + 1$
}
\end{algorithm}


\section{Experiments}

\begin{table*}[h]
\centering
\resizebox{0.85\linewidth}{!}{
\begin{tabular}{ccc|cccc|cccc}
\toprule
\multirow{2}{*}{\textbf{Script}} & \multirow{2}{*}{\textbf{Evaluation}} & \multirow{2}{*}{\textbf{\#QA}} 
& \multicolumn{4}{c}{\textbf{GPT-3.5}} 
& \multicolumn{4}{c}{\textbf{Qwen2.5 32B}}\\
\cmidrule(lr){4-7}\cmidrule(lr){8-11}
 &  &  & \textbf{Werewolf} & \textbf{O-CoT} & \textbf{ThinkThrice} & \textbf{\agentname} 
 & \textbf{Werewolf} & \textbf{O-CoT} & \textbf{ThinkThrice} & \textbf{\agentname} \\

\midrule
\multirow{5}{*}{\shortstack[c]{\emph{Death Wears White}\\ \emph{(9 players, 1 victim)}}}
&  Win Rate &  -  & $.000_{\pm.000}$ & $.000_{\pm.000}$ & $.000_{\pm.000}$ & $.000_{\pm.000}$ & $.000_{\pm.000}$ & $.000_{\pm.000}$ & $.000_{\pm.000}$ & $.000_{\pm.000}$ \\
&   Objective  &  $ 10 $   & $.033_{\pm.047}$ & $.067_{\pm.047}$ & $.067_{\pm.047}$ & $.033_{\pm.047}$ & $.033_{\pm.047}$ & $.067_{\pm.047}$ & $.267_{\pm.094}$ & $.200_{\pm.082}$ \\
&   Reasoning  & $ 102$     & $.258_{\pm.024}$ & $.324_{\pm.014}$ & $.327_{\pm.012}$ & $.356_{\pm.030}$ & $.458_{\pm.009}$ & $.467_{\pm.018}$ & $.441_{\pm.021}$ & $.539_{\pm.008}$ \\
&   Relations  & $ 72 $   & $.421_{\pm.024}$ & $.458_{\pm.041}$ & $.384_{\pm.026}$ & $.435_{\pm.017}$ & $.699_{\pm.029}$ & $.764_{\pm.030}$ & $.764_{\pm.023}$ & $.741_{\pm.036}$ \\
\midrule
\multirow{5}{*}{\shortstack[c]{\emph{Ghost Revenge}\\ \emph{(7 players, 3 victims)}}}
&  Win Rate  &  $-$ & $.222_{\pm.157}$ & $.111_{\pm.157}$ & $.000_{\pm.000}$ & $.333_{\pm.000}$ & $.333_{\pm.000}$ & $.000_{\pm.000}$ & $.333_{\pm.000}$ & $.333_{\pm.000}$ \\
&   Objective   &  $ 19 $  & $.193_{\pm.066}$ & $.158_{\pm.043}$ & $.193_{\pm.025}$ & $.333_{\pm.066}$ & $.333_{\pm.066}$ & $.088_{\pm.025}$ & $.316_{\pm.043}$ & $.351_{\pm.066}$ \\
&   Reasoning   &  $ 152 $   & $.307_{\pm.008}$ & $.322_{\pm.019}$ & $.377_{\pm.022}$ & $.353_{\pm.016}$ & $.452_{\pm.012}$ & $.461_{\pm.027}$ & $.458_{\pm.014}$ & $.511_{\pm.017}$ \\
&   Relations  & $ 69 $  & $.314_{\pm.018}$ & $.295_{\pm.018}$ & $.353_{\pm.025}$ & $.353_{\pm.018}$ & $.671_{\pm.030}$ & $.599_{\pm.018}$ & $.652_{\pm.012}$ & $.686_{\pm.058}$ \\
\midrule
\multirow{5}{*}{\shortstack[c]{\emph{Danshui Villa}\\ \emph{(7 players, 2 victims)}}}
&  Win Rate & $-$ & $.000_{\pm.000}$ & $.000_{\pm.000}$ & $.000_{\pm.000}$ & $.000_{\pm.000}$ & $.333_{\pm.236}$ & $.500_{\pm.000}$ & $.667_{\pm.236}$ & $.500_{\pm.000}$ \\
&   Objective    &  $ 12 $ & $.083_{\pm.000}$ & $.111_{\pm.039}$ & $.194_{\pm.039}$ & $.111_{\pm.079}$ & $.444_{\pm.079}$ & $.361_{\pm.039}$ & $.528_{\pm.171}$ & $.472_{\pm.039}$ \\
&   Reasoning & $ 128 $ & $.286_{\pm.027}$ & $.286_{\pm.019}$ & $.310_{\pm.004}$ & $.286_{\pm.004}$ & $.372_{\pm.004}$ & $.424_{\pm.029}$ & $.398_{\pm.023}$ & $.440_{\pm.004}$ \\
&   Relations & $ 63 $  & $.312_{\pm.027}$ & $.365_{\pm.045}$ & $.259_{\pm.015}$ & $.296_{\pm.075}$ & $.577_{\pm.046}$ & $.614_{\pm.020}$ & $.571_{\pm.013}$ & $.571_{\pm.013}$ \\
\midrule
\multirow{5}{*}{\shortstack[c]{\emph{Unfinished Love }\\ \emph{(7 players, 2 victims)}}}
&  Win Rate  &  $-$ & $.000_{\pm.000}$ & $.000_{\pm.000}$ & $.000_{\pm.000}$ & $.500_{\pm.000}$ & $.167_{\pm.236}$ & $.333_{\pm.236}$ & $.167_{\pm.236}$ & $.500_{\pm.000}$ \\
&   Objective & $ 12 $   & $.083_{\pm.118}$ & $.028_{\pm.039}$ & $.000_{\pm.000}$ & $.333_{\pm.068}$ & $.222_{\pm.079}$ & $.250_{\pm.068}$ & $.194_{\pm.039}$ & $.361_{\pm.039}$ \\
&   Reasoning & $ 61 $   & $.443_{\pm.023}$ & $.426_{\pm.035}$ & $.481_{\pm.039}$ & $.536_{\pm.028}$ & $.530_{\pm.008}$ & $.557_{\pm.027}$ & $.601_{\pm.015}$ & $.590_{\pm.027}$ \\
&   Relations &  $ 72 $    & $.472_{\pm.034}$ & $.519_{\pm.036}$ & $.514_{\pm.020}$ & $.560_{\pm.007}$ & $.764_{\pm.023}$ & $.745_{\pm.007}$ & $.782_{\pm.013}$ & $.778_{\pm.011}$ \\
\midrule
\multirow{5}{*}{\shortstack[c]{\emph{Cruise Incident}\\ \emph{(5 players, 1 victim)}}}
&  Win Rate &  $-$  & $.667_{\pm.471}$ & $1.000_{\pm.000}$ & $.667_{\pm.471}$ & $1.000_{\pm.000}$ & $.000_{\pm.000}$ & $.000_{\pm.000}$ & $.000_{\pm.000}$ & $.333_{\pm.471}$ \\
&   Objective & $ 4 $ & $.417_{\pm.118}$ & $.500_{\pm.000}$ & $.583_{\pm.312}$ & $.667_{\pm.236}$ & $.000_{\pm.000}$ & $.000_{\pm.000}$ & $.000_{\pm.000}$ & $.250_{\pm.204}$ \\
&   Reasoning & $ 24 $  & $.458_{\pm.068}$ & $.444_{\pm.086}$ & $.458_{\pm.059}$ & $.528_{\pm.052}$ & $.639_{\pm.020}$ & $.708_{\pm.059}$ & $.667_{\pm.000}$ & $.778_{\pm.071}$ \\
&   Relations & $ 30 $ & $.367_{\pm.082}$ & $.422_{\pm.042}$ & $.411_{\pm.016}$ & $.422_{\pm.016}$ & $.833_{\pm.047}$ & $.789_{\pm.016}$ & $.700_{\pm.027}$ & $.711_{\pm.016}$ \\
\midrule
\multirow{5}{*}{\shortstack[c]{\emph{Sin}\\ \emph{(4 players, 1 victim)}}}
&  Win Rate  &  $-$ & $.333_{\pm.471}$ & $.000_{\pm.000}$ & $.000_{\pm.000}$ & $.667_{\pm.471}$ & $.000_{\pm.000}$ & $.000_{\pm.000}$ & $.000_{\pm.000}$ & $1.000_{\pm.000}$ \\
&   Objective & $ 3 $   & $.333_{\pm.272}$ & $.000_{\pm.000}$ & $.000_{\pm.000}$ & $.444_{\pm.314}$ & $.000_{\pm.000}$ & $.000_{\pm.000}$ & $.000_{\pm.000}$ & $1.000_{\pm.000}$ \\
&   Reasoning & $ 20 $ & $.650_{\pm.108}$ & $.467_{\pm.024}$ & $.533_{\pm.047}$ & $.550_{\pm.041}$ & $.717_{\pm.047}$ & $.550_{\pm.000}$ & $.633_{\pm.047}$ & $.700_{\pm.041}$ \\
&   Relations & $ 21 $ & $.333_{\pm.067}$ & $.571_{\pm.067}$ & $.413_{\pm.090}$ & $.492_{\pm.022}$ & $.730_{\pm.059}$ & $.889_{\pm.022}$ & $.698_{\pm.045}$ & $.794_{\pm.022}$ \\
\midrule
\multirow{5}{*}{\shortstack[c]{\emph{Deadly Fountain }\\ \emph{(4 players, 1 victim)}}}
&  Win Rate  &  $-$ & $.000_{\pm.000}$ & $.000_{\pm.000}$ & $.000_{\pm.000}$ & $.000_{\pm.000}$ & $.000_{\pm.000}$ & $.000_{\pm.000}$ & $.000_{\pm.000}$ & $.000_{\pm.000}$ \\
&   Objective  &$ 3 $  & $.000_{\pm.000}$ & $.000_{\pm.000}$ & $.000_{\pm.000}$ & $.000_{\pm.000}$ & $.000_{\pm.000}$ & $.000_{\pm.000}$ & $.000_{\pm.000}$ & $.222_{\pm.157}$ \\
&   Reasoning & $ 21 $ & $.381_{\pm.000}$ & $.444_{\pm.098}$ & $.508_{\pm.022}$ & $.587_{\pm.022}$ & $.540_{\pm.022}$ & $.556_{\pm.022}$ & $.508_{\pm.045}$ & $.587_{\pm.022}$ \\
&   Relations & $ 12 $  & $.250_{\pm.068}$ & $.194_{\pm.039}$ & $.222_{\pm.039}$ & $.333_{\pm.068}$ & $.583_{\pm.118}$ & $.667_{\pm.136}$ & $.667_{\pm.068}$ & $.667_{\pm.136}$ \\
\midrule
\multirow{5}{*}{\shortstack[c]{\emph{Unbelievable Incident}\\ \emph{(5 players, 1 victim)}}}
&  Win Rate  &  $-$ & $.000_{\pm.000}$ & $.000_{\pm.000}$ & $.000_{\pm.000}$ & $.000_{\pm.000}$ & $.000_{\pm.000}$ & $.000_{\pm.000}$ & $.000_{\pm.000}$ & $.000_{\pm.000}$ \\
&   Objective &$ 4 $ & $.083_{\pm.118}$ & $.000_{\pm.000}$ & $.000_{\pm.000}$ & $.083_{\pm.118}$ & $.000_{\pm.000}$ & $.083_{\pm.118}$ & $.083_{\pm.118}$ & $.000_{\pm.000}$ \\
&   Reasoning & $ 24 $  & $.431_{\pm.052}$ & $.278_{\pm.071}$ & $.194_{\pm.071}$ & $.472_{\pm.020}$ & $.292_{\pm.000}$ & $.319_{\pm.020}$ & $.278_{\pm.020}$ & $.389_{\pm.020}$ \\
&   Relations & $ 15 $   & $.622_{\pm.083}$ & $.733_{\pm.109}$ & $.289_{\pm.126}$ & $.533_{\pm.054}$ & $.756_{\pm.063}$ & $.822_{\pm.031}$ & $.733_{\pm.000}$ & $.822_{\pm.031}$ \\
\midrule
\multirow{5}{*}{\shortstack[c]{\emph{Desperate Sunshine}\\ \emph{(4 players, 1 victim)}}}
&  Win Rate  &  $-$ & $.000_{\pm.000}$ & $.000_{\pm.000}$ & $.667_{\pm.471}$ & $1.000_{\pm.000}$ & $.333_{\pm.471}$ & $.000_{\pm.000}$ & $.667_{\pm.471}$ & $.000_{\pm.000}$ \\
&   Objective  & $ 3 $& $.333_{\pm.000}$ & $.111_{\pm.157}$ & $.556_{\pm.157}$ & $.778_{\pm.157}$ & $.333_{\pm.272}$ & $.333_{\pm.000}$ & $.556_{\pm.157}$ & $.333_{\pm.000}$ \\
&   Reasoning & $ 18 $  & $.537_{\pm.026}$ & $.630_{\pm.094}$ & $.759_{\pm.052}$ & $.741_{\pm.069}$ & $.741_{\pm.052}$ & $.778_{\pm.045}$ & $.741_{\pm.026}$ & $.778_{\pm.000}$ \\
&   Relations  & $ 36 $  & $.491_{\pm.094}$ & $.574_{\pm.035}$ & $.491_{\pm.047}$ & $.556_{\pm.068}$ & $.778_{\pm.023}$ & $.815_{\pm.013}$ & $.787_{\pm.013}$ & $.787_{\pm.035}$ \\
\midrule
\multirow{5}{*}{\shortstack[c]{\emph{Riverside Inn}\\ \emph{(4 players, 1 victim)}}}
&  Win Rate  &  $-$ & $.083_{\pm.118}$ & $.000_{\pm.000}$ & $.083_{\pm.118}$ & $.250_{\pm.204}$ & $.083_{\pm.118}$ & $.000_{\pm.000}$ & $.000_{\pm.000}$ & $.083_{\pm.118}$ \\
&   Objective & $ 3 $ & $.117_{\pm.024}$ & $.117_{\pm.062}$ & $.117_{\pm.094}$ & $.383_{\pm.062}$ & $.167_{\pm.047}$ & $.167_{\pm.024}$ & $.133_{\pm.062}$ & $.250_{\pm.041}$ \\
&   Reasoning & $ 18 $  & $.248_{\pm.013}$ & $.312_{\pm.022}$ & $.281_{\pm.011}$ & $.373_{\pm.017}$ & $.373_{\pm.016}$ & $.339_{\pm.015}$ & $.394_{\pm.007}$ & $.413_{\pm.020}$ \\
&   Relations & $ 18 $ & $.531_{\pm.014}$ & $.541_{\pm.038}$ & $.483_{\pm.030}$ & $.589_{\pm.071}$ & $.700_{\pm.025}$ & $.705_{\pm.025}$ & $.647_{\pm.018}$ & $.681_{\pm.059}$ \\
\midrule
\multirow{5}{*}{\shortstack[c]{\emph{Solitary Boat Firefly}\\ \emph{(6 players, 4 victims)}}}
&  Win Rate  &  $-$ & $.222_{\pm.157}$ & $.000_{\pm.000}$ & $.000_{\pm.000}$ & $.000_{\pm.000}$ & $.333_{\pm.000}$ & $.111_{\pm.157}$ & $.111_{\pm.157}$ & $.556_{\pm.157}$ \\
&   Objective  & $ 20 $ & $.250_{\pm.090}$ & $.181_{\pm.039}$ & $.181_{\pm.039}$ & $.250_{\pm.034}$ & $.389_{\pm.079}$ & $.236_{\pm.098}$ & $.264_{\pm.109}$ & $.472_{\pm.052}$ \\
&   Reasoning  &  $ 109 $ & $.409_{\pm.004}$ & $.369_{\pm.044}$ & $.453_{\pm.008}$ & $.539_{\pm.027}$ & $.553_{\pm.013}$ & $.537_{\pm.027}$ & $.602_{\pm.018}$ & $.618_{\pm.018}$ \\
&   Relations &  $ 69 $ & $.473_{\pm.028}$ & $.568_{\pm.048}$ & $.439_{\pm.033}$ & $.542_{\pm.027}$ & $.758_{\pm.030}$ & $.788_{\pm.019}$ & $.739_{\pm.009}$ & $.780_{\pm.014}$ \\
\midrule
\multirow{5}{*}{\shortstack[c]{\emph{Manna}\\ \emph{(6 players, 3 victims)}}}
&  Win Rate &  $-$ & $.000_{\pm.000}$ & $.000_{\pm.000}$ & $.667_{\pm.471}$ & $.333_{\pm.471}$ & $.000_{\pm.000}$ & $.000_{\pm.000}$ & $.000_{\pm.000}$ & $1.000_{\pm.000}$ \\
&   Objective  & $ 24 $  & $.111_{\pm.157}$ & $.000_{\pm.000}$ & $.556_{\pm.157}$ & $.444_{\pm.157}$ & $.000_{\pm.000}$ & $.333_{\pm.000}$ & $.111_{\pm.157}$ & $1.000_{\pm.000}$ \\
&   Reasoning  &  $ 123 $ & $.463_{\pm.052}$ & $.426_{\pm.026}$ & $.593_{\pm.026}$ & $.648_{\pm.069}$ & $.648_{\pm.026}$ & $.759_{\pm.026}$ & $.685_{\pm.026}$ & $.815_{\pm.026}$ \\
&   Relations  & $ 88 $ & $.444_{\pm.045}$ & $.407_{\pm.069}$ & $.333_{\pm.045}$ & $.444_{\pm.045}$ & $.778_{\pm.045}$ & $.741_{\pm.069}$ & $.741_{\pm.069}$ & $.815_{\pm.052}$ \\
\midrule
\multirow{5}{*}{\shortstack[c]{Overall}}
&  Win Rate & -  & $.127_{\pm.059}$ & $.063_{\pm.022}$ & $.111_{\pm.045}$ & $\mathbf{.222}_{\pm.081}$ & $.175_{\pm.081}$ & $.095_{\pm.039}$ & $.175_{\pm.059}$ & $\mathbf{.349}^*_{\pm.045}$ \\
&    Objective & $  117  $ & $.160_{\pm.020}$ & $.123_{\pm.008}$ & $.162_{\pm.024}$ & $\mathbf{.288}^*_{\pm.021}$ & $.242_{\pm.049}$ & $.179_{\pm.030}$ & $.245_{\pm.059}$ & $\mathbf{.373}^*_{\pm.031}$ \\
&    Reasoning & $  800  $  & $.343_{\pm.006}$ & $.349_{\pm.012}$ & $.384_{\pm.002}$ & $\mathbf{.423}^*_{\pm.009}$ & $.471_{\pm.001}$ & $.480_{\pm.015}$ & $.489_{\pm.003}$ & $\mathbf{.536}^*_{\pm.011}$ \\
&    Relations  &  $  565  $ & $.425_{\pm.008}$ & $\mathbf{.473}_{\pm.010}$ & $.405_{\pm.006}$ & $.471_{\pm.009}$ & $.714_{\pm.006}$ & $\mathbf{.729}_{\pm.009}$ & $.705_{\pm.003}$ & $.725_{\pm.011}$ \\
&    Overall   & $  1482  $ & $.324_{\pm.004}$ & $.329_{\pm.008}$ & $.347_{\pm.006}$ & $\mathbf{.407}^*_{\pm.006}$ & $.472_{\pm.010}$ & $.469_{\pm.016}$ & $.482_{\pm.013}$ & $\mathbf{.540}^*_{\pm.013}$ \\
\bottomrule

\end{tabular}}
\caption{Results of Agent-vs-Agent Evaluation. An ``\textasteriskcentered'' indicates statistical significance under the two-sample t-test with a significance level of $\alpha = 0.05$, comparing it with the second-best model. 
}
\label{tab:main-results-short}
\end{table*}

\subsection{Experimental Setup}
We conducted experiments with GPT-3.5 (gpt-35-turbo-16k) and Qwen2.5-32B-Instruct~\citep{yang2024qwen2technicalreport} for conversation, and the GPT Embedding Model (text-embedding-ada-002) for memory retrieval.
To minimise randomness, we conducted the evaluation experiments 3 times and reported the average and standard deviation.
In this paper, we focus specifically on the strategy for the good camp. The murderer agent's questioning strategy closely mirrors that of other agents, with an added prompt explicitly instructing it to act as if it is not the murderer while questioning others.


\subsection{Evaluation Metrics}
For evaluation, we adopt the following metrics:

\paragraph{Win Rate:} Every player, including the murderer, can vote, and the player with the majority number of votes ($\geq 50\%$) is eliminated. Murderers always vote for others instead of themselves. If the true murderer is voted out, the game is considered a victory for the identifying players.
\paragraph{Question Accuracy:} We report agents’ accuracy on three question types: Objective, Reasoning, and Relations, along with an Overall score that aggregates performance across all types. This metric is used in both Agent-vs-Agent and Human-vs-Agent evaluations to measure agents' ability to answer inferential questions correctly.

\subsection{Baselines}
For baselines, we compare our approach with other multi-agent algorithms designed for multiplayer deduction games.  Although some methods were not designed for MMGs, they are the most relevant and adaptable frameworks in this relatively new area of LLM-based multi-agent games. 

(1) \emph{Werewolf} \citep{werewolf-xu2023} is another multi-agent game, where players identify werewolves through group discussion. Questions are chosen from a role-specific predefined list to facilitate game progression, alongside questions generated based on the current scenario. We adapt its pre-set game instructions and role-specific information to MMGs settings.
(2) \emph{Objective-Guided Chain of Thought} (O-CoT) \citep{generative-agent-park2023, narrativeplay-zhao2024}. Agents think, reflect, and choose who and what to ask based on their objectives. We use the framework from previous works, only replacing the agents' objectives with those set in MMGs.
(3) \emph{ThinkThrice} \citep{murdergame-wu2024}. Designed for MMGs, agents craft questions from retrieved memory and the current scenario. 
(4) \emph{Personal Perspective} (PP). For a more comprehensive comparison, we also assess the performance of agents who do not actively participate in the game but make their final decisions only based on their script.
(5) \emph{Omniscient Perspective} (OP). Agents do not actively participate but make their final decisions based on all agents' scripts.

To account for the zero-sum nature of MMGs, where it would be hard to tell if agents can identify the murderer due to their good performance or the poor performance of their competitor, we fix the murderer's framework in all experiments (including baselines and \agentname). 
The implementation details, model comparison, information on sensors, method to calculate the Overall, and prompts are provided in Appendix \ref{appendix:Implementation}.

\subsection{Agent-vs-Agent Evaluation}

\begin{figure*}[th!]
    \centering
    \includegraphics[width=\linewidth]{evaluation_scores_hd.png}
    \caption{Comparison of the performance of agents with other multi-agent algorithms designed for multiplayer deduction games. The Personal Perspective (PP) baseline represents the starting point for searching, where agents rely solely on their own knowledge. The Omniscient Perspective (OP) measures performance when agents have full access to all other agents’ scripts, representing the ideal search endpoint.  
    }
    \label{fig:performance}
\end{figure*}

In this section, we evaluate agent performance in an Agent-vs-Agent setting to compare their decision-making processes against competitive baselines.
Figure~\ref{fig:performance} presents the average of agents' performance across 12 unique games of varying complexity and settings as detailed in our results table (Table~\ref{tab:main-results-short}). 
\agentname \  shows superior performance compared to other baselines across all evaluation questions, demonstrating its enhanced understanding of the search space through interactions with other agents. Notably, \agentname \  significantly outperforms others in Objective Questions. The OP setting, where agents have access to the scripts of all agents without interacting with them, generally yields better performance compared to all other methodologies. It represents the ideal \emph{``search''} endpoint. However, in practice, its effectiveness is often limited by the deductive capabilities of the underlying base model, which reflects the upper limit of the \emph{``approximate''} ability of the given base model. This showcases the \emph{``approximate''} ability of \agentname \  in refining information and dynamically narrowing down the search domain to achieve the target objective. Since the PP and OP settings do not involve active participation in the game, we consider them as indicators of the \emph{starting point} and \emph{endpoint} that can be achieved through search. 

The results demonstrate that \agentname \  outperforms all baseline agents in reasoning and relations, showcasing its superior \emph{``search''} ability. The larger performance gap in objective questions for other baselines reveals a critical limitation: an inability to effectively utilise the collected information to reach correct conclusions or achieve game-specific goals. This highlights the importance of not only gathering relevant information but also efficiently processing and applying it to attain the desired objectives.
Detailed dialogue history and evaluation records are in the Git link provided previously. 

\subsection{Agent-vs-Human Evaluation}

\begin{figure*}[t]
    \centering
    \includegraphics[width=\linewidth]{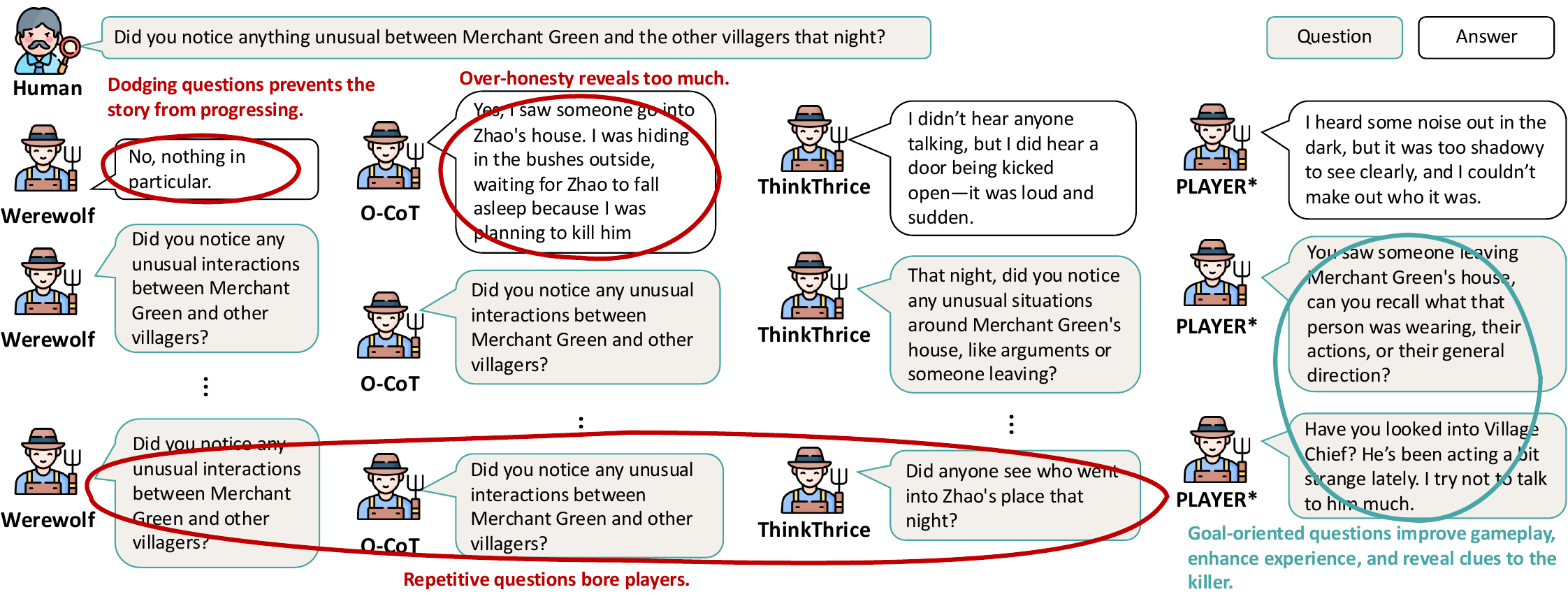}
    \caption{\textbf{Comparison of Dialogue Strategies.} \agentname \ significantly enhances story progression by eliciting key clues (clothing, movement, direction), guiding the investigation. It demonstrates superior questioning by targeting specific details, leading to richer responses. In contrast, \textbf{Werewolf} provides minimal advancement, \textbf{ThinkThrice} adds vague auditory clues, and \textbf{O-CoT} being overly honest leads to revealing things that should be kept secret. Overall, \agentname \ outperforms others by  designing dialogue for better narrative engagement.}
    \label{agent-vs-human-examples}
\end{figure*}

In the Agent-vs-Human Evaluation, we identified significant limitations of traditional evaluation methods used in agent for games, which were primarily designed for Agent-vs-Agent Evaluation and focused solely on achieving victory. While effective in assessing agents’ strategic and reasoning abilities, these methods fall short when considering human players as the end-users of the system. For instance, previous approaches often resulted in repetitive and meaningless conversations, which human players found unengaging and boring. 
To illustrate these limitations more clearly, we provide specific examples in Figure~\ref{agent-vs-human-examples}, carefully drawn from real dialogues that occurred during games involving human participants.


To address this, we extended our evaluation by incorporating a player-centric perspective. In addition to the performance metrics used in Agent-vs-Agent Evaluation, we distributed a survey to human participants to capture their gameplay experience and satisfaction levels. This allowed us to assess \agentname's suitability not only as a strategic agent but also as a companion for human players in interactive settings. The metrics include:

\noindent \textbf{Story Advancement:} Measures the agent’s effectiveness in gathering relevant information and advancing the game’s plot in a meaningful way. Higher scores indicate that the interactions were productive and kept the game engaging.

 \noindent \textbf{Question Quality:} Evaluates the relevance and depth of the questions posed by the agent. This metric ensures that the questions contribute to the narrative and help uncover critical information, avoiding redundancy or triviality.

\noindent \textbf{Response Quality:} Assesses the accuracy and informativeness of the agent’s responses. Higher scores indicate that the agent provided clear and valuable information to the human players.

\noindent \textbf{Response Speed:} Represents how quickly the agent interacts, ensuring timely response for a smooth and engaging gameplay experience.

\noindent \textbf{Role Immersion:} Assesses how well the agent embodies its character’s personality, emotions, and background to create a believable experience.

\begin{table}[h]
\centering
\resizebox{\linewidth}{!}{
\begin{tabular}{ccccc}
\toprule
\textbf{Evaluation} & \textbf{Werewolf} & \textbf{O-CoT} & \textbf{ThinkThrice} & \textbf{\agentname \ } \\
\midrule
Win Rate  & $.111_{\pm.157}$ & $.333_{\pm.132}$ & $.333_{\pm.048}$ & \bm{$.667_{\pm.085}$} \\
Objective & $.222_{\pm.091}$ & $.333_{\pm.045}$ & $.296_{\pm.052}$ &\bm{ $.556_{\pm.091}$} \\
Reasoning & $.678_{\pm.028}$ & $.644_{\pm.028}$ & $.684_{\pm.008}$ & \bm{$.723_{\pm.035}$ }\\
Relations & $.745_{\pm.028}$ & $.797_{\pm.040}$ & $.745_{\pm.028}$ &\bm{ $.843_{\pm.042}$} \\
Overall   & $.608_{\pm.006}$ & $.619_{\pm.023}$ & $.625_{\pm.009}$ & \bm{$.717_{\pm.038}$ }\\
\bottomrule
\end{tabular}}
\caption{Performance comparison of different algorithms against human players.}
\label{tab:human-results}
\end{table}

\begin{figure}[h!]
    \centering
    \includegraphics[width=0.9\linewidth]{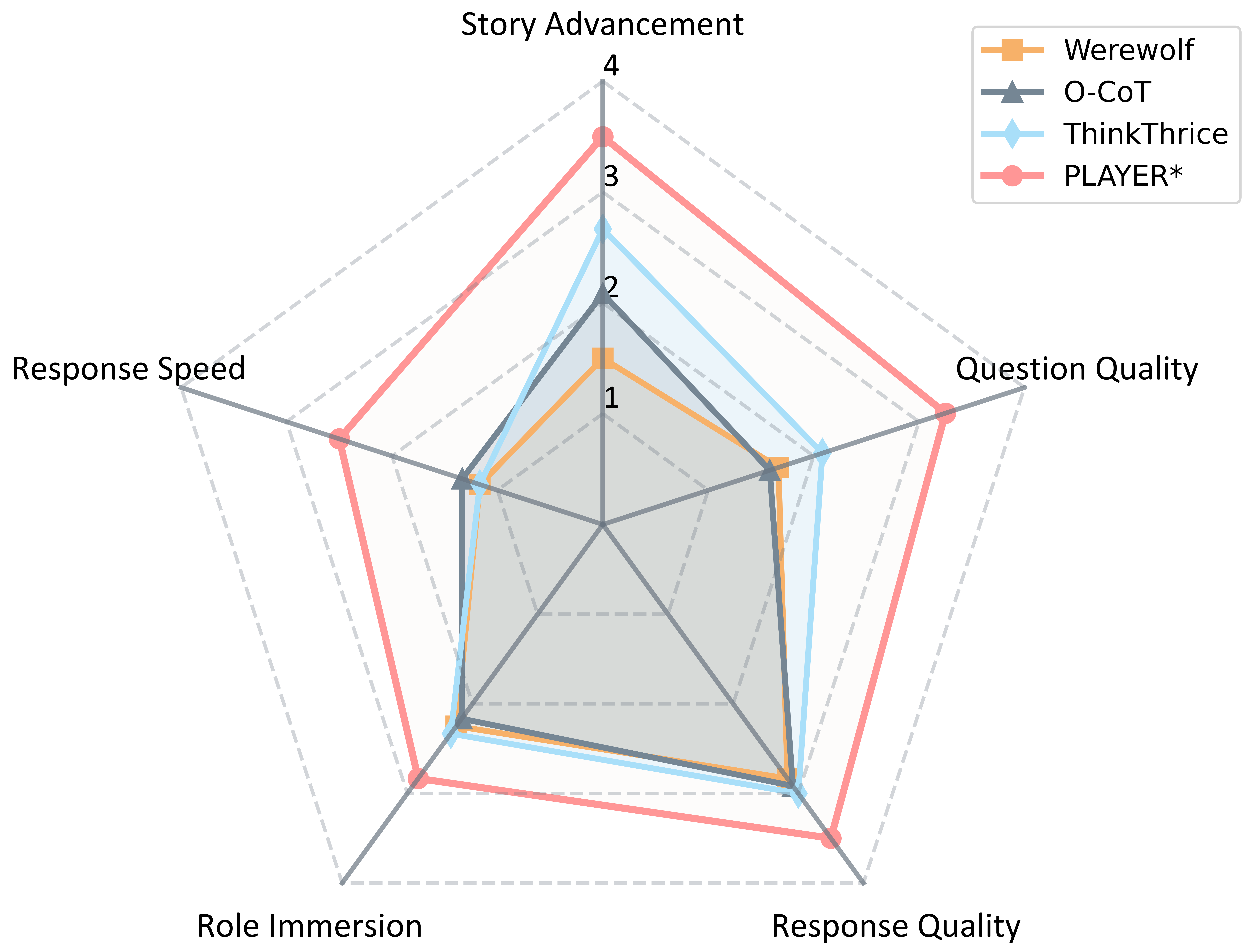}
    \caption{ Results of Agent-vs-Human Evaluation using
Human-Centric metrics. The detailed data can be found in Table~\ref{tab:human-survey}. }
    \label{fig:human_eva_radar}
\end{figure}

All scores are based on discrete 1-5 scales, where higher scores are better. Details about the setups (e.g., recruitment of players, gameplay conditions, and tasks) and scoring criteria can be found in Appendix~\ref{human-setup}. 



As shown in Table~\ref{tab:human-results} (results against human players) and Figure~\ref{fig:human_eva_radar} (Human-Centric metrics), \agentname\ not only clearly surpassed the baseline models in strategic and reasoning-oriented evaluation but also demonstrated superior performance in key human-centric dimensions. These results strongly indicate that \agentname\ effectively balances tactical prowess with narrative engagement, delivering more satisfying and immersive gameplay experiences for human participants.

\subsection{Efficiency and Cost Analysis}

\begin{figure}[h!]
    \centering
    \includegraphics[width=\linewidth]{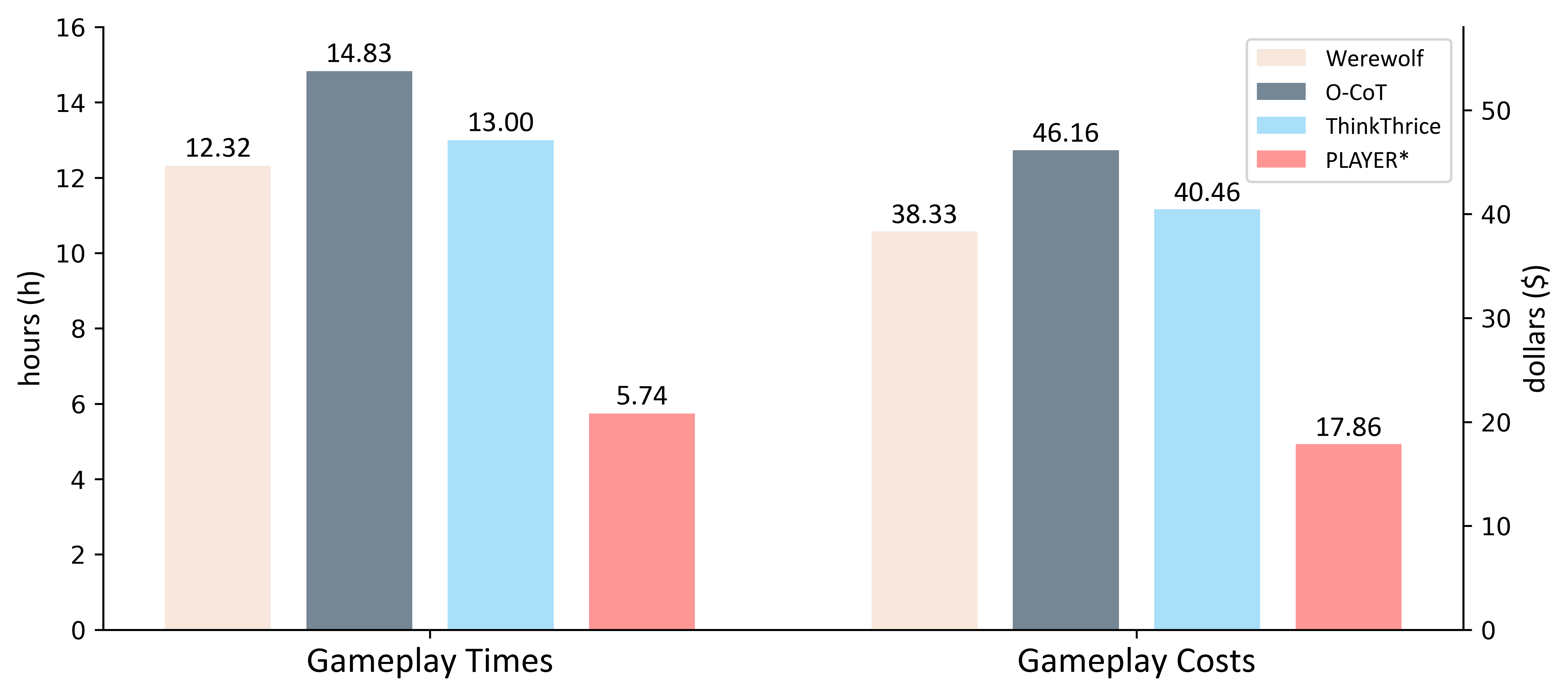}
    \caption{ Comparison of time (hours) and costs (\$) for calling OpenAI API across multi-agent algorithms in MMG settings. 
    }
    \label{fig:cost}
\end{figure}

As shown in Figure~\ref{fig:cost}, we delve into the efficiency and cost analysis across various methodologies for agent interaction in MMGs (detailed in Table~\ref{tab:cost-results}). 
The costs are presented in actual monetary values (US Dollars) associated with the use of the Azure API service, providing a direct and transparent measure of the computational expense.\footnote{Billing method details are available on the website \url{https://azure.microsoft.com/en-gb/pricing/details/cognitive-services/openai-service/.}}

Werewolf runs efficiently due to its use of preset instructions, while O-CoT is the most costly, requiring step-by-step reasoning. \agentname\ stands out for its cost-efficiency and strong performance. Its pruning strategy reduces unnecessary API calls by focusing on the most suspicious characters, lowering both time and cost. This efficiency holds across all scripts. 

\subsection{Ablation Studies} \label{ablation study}

As shown in Table~\ref{tab:asymmetric_incompatible_res}, we can observe that the two main modules ``Search via state matching" and ``Approximate with pruner" in our algorithm both reduced the cost and improved the performance as expected. \agentname \  equipped with both modules gives the best performance.

\begin{table} [h]
    \centering
     \resizebox{\columnwidth}{!}{
    \begin{tabular}{c c c c}
    \toprule

    & Score & Gameplay Costs (\$) & Gameplay Times (h)\\
    \midrule

    PLAYER* \  & 0.407  &  17.861    &  5.740  \\

    w/o Pruner & 0.396  &  28.731    & 9.932 \\
        w/o Sensor &0.379  &  15.057    & 4.457 \\
    w/o Sensor \& Pruner & 0.357   &19.144   & 6.577  \\
    \bottomrule
    \end{tabular}
    }
    \caption{Ablation study of removing the Pruner and Sensor on performance, gameplay costs, and times.}
    \label{tab:asymmetric_incompatible_res}
    \vspace{-1em}
\end{table}

\paragraph{Sensor Selection}

\begin{figure}[h!]
    \centering
    \includegraphics[width=\linewidth]{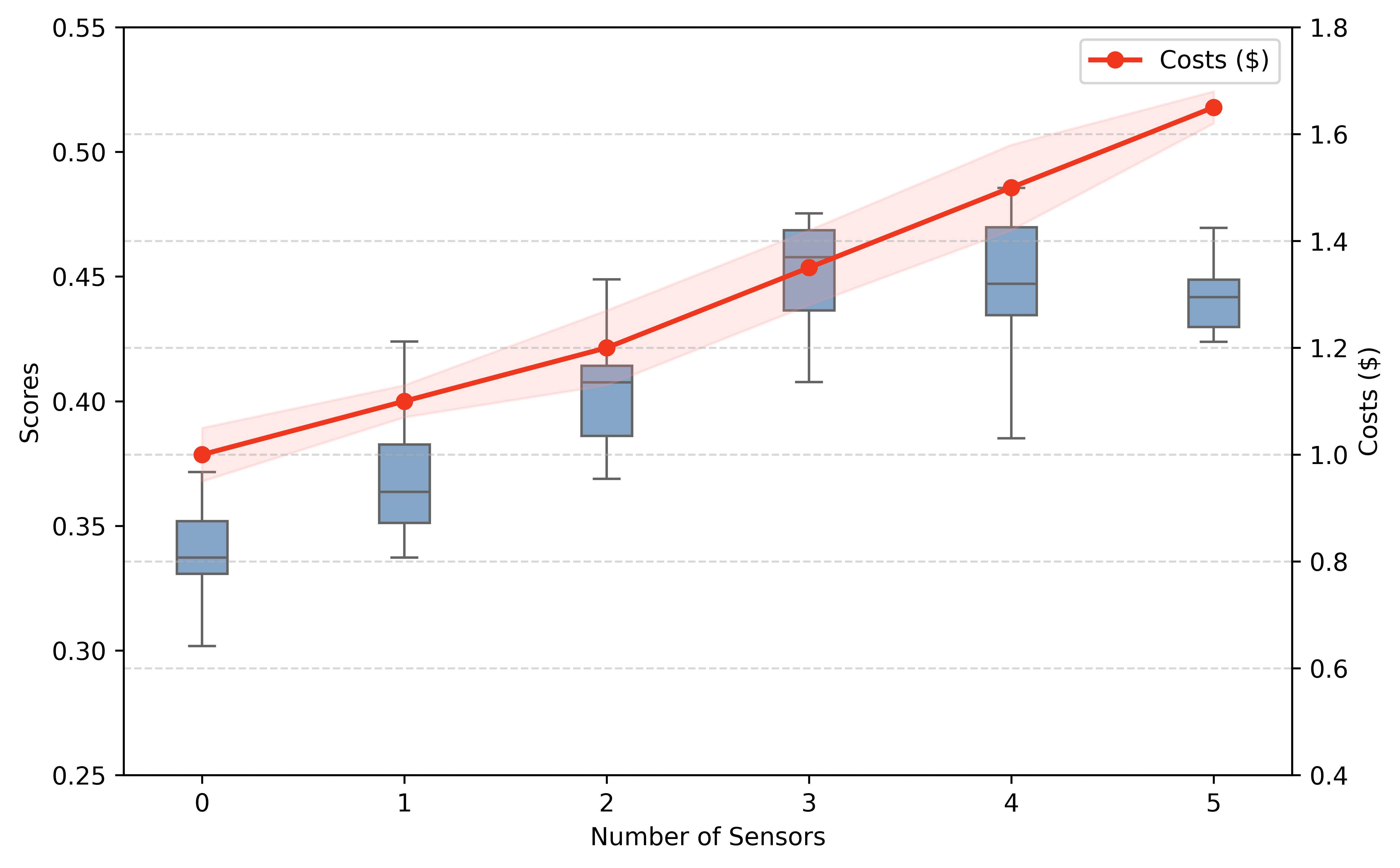}
    \caption{Compare the number of sensors, scores, and costs. The chart displays the distribution of scores across different sensor combinations using box plots, with the red line representing the corresponding costs and the shaded area indicating variability.
}
    \label{fig:sensor_nunber}
\end{figure}

In practice, sensors may exhibit non-linear relationships, which necessitates careful evaluation to avoid redundancy. So we assess the unique contributions of each new sensor, ensuring that every selected sensor introduces meaningful and distinct information. For MMGs, we initialise 5 sensors with domain knowledge: 
\emph{Emotion} (an agent’s disposition toward another character, reflecting their willingness to help or uncover their issues), \emph{Motivation} (whether the character has a potential motive from the agent’s perspective), \emph{Opportunity Assessment} (whether the character had the chance to commit the crime), \emph{Evidence} (whether there is direct evidence linking the character to the crime scene), and \emph{Background} (whether there is a history of conflict, rivalry, or enmity with others).
As shown in Figure~\ref{fig:sensor_nunber}, we evaluate the performance gain provided by each sensor. The benefits diminish as additional sensors contribute less new information, while the negative impact from longer input sequences grows. Moreover, adding more sensors increases computational costs. To balance sensor effectiveness with task complexity, we selected three sensors. The final combination—Emotion, Motivation, and Opportunity Assessment—offered the best performance in our experiments. The experiment was run with different combinations of sensors. For example, $\binom{5}{2}$ runs to cover all possible combinations of sensors when the number of sensors is $2$.

\begin{figure}[t]
    \centering
    \includegraphics[width=\linewidth]{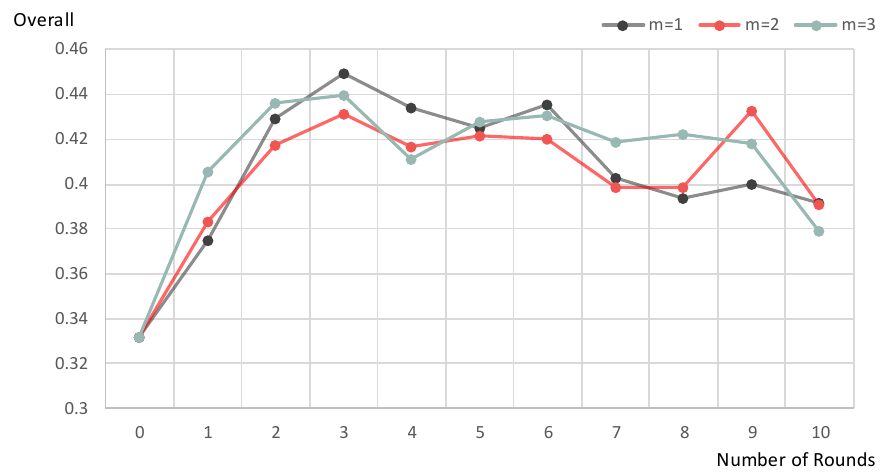}
    \caption{Comparison of agents' behaviour across different numbers of rounds, where each agent can ask a specific number of questions (denoted as $m$).}
    \label{fig:ablation-studies}
\end{figure}

\paragraph{Optimal Number of Rounds and \agentname \  plannings}
We compared agents' behavior across different numbers of rounds and with varying numbers of questions agents can ask per round, as shown in Figure~\ref{fig:ablation-studies}.
Performance peaks around round 3, after which it shows variability, with some rounds experiencing slight declines or plateauing in scores despite more rounds or questions. This indicates that after a rapid initial learning or adaptation phase, where agents effectively use additional questions to enhance their understanding and strategies, the value of IGed from conversations tends to converge. These results also provide empirical support for the assumption made in our methodology that the more inquiries we pose to an agent, the expected reward associated with questioning the same agent decreases.
For the main results we report, we use the original setting for the number of rounds in MMG, which is 3, and based on the outcomes of the ablation studies, we chose the most effective number of questions to ask per round, which is $1$.

\section{Related Works}
\paragraph{Multi-Agent Interaction}
Multi-agent reinforcement learning marking significant progress in complex games \citep{marl-lanctot2017, marl-perolat2022, marl-bakhtin2023}. However, these methods often require extensive computational resources and lack linguistic communication capabilities. With the emergence of LLMs, there's a shift of focus towards improving multi-agent language communication, evidenced by advancements in various games and scenarios, such as werewolf \citep{werewolf-xu2023}, avalon \citep{avalon-wang2023, avalon-shi2024}, interactive narrative \citep{narrativeplay-zhao2024}, MMGs \citep{murdergame-wu2024}, and survival games \citep{ metacognition-toy2024}. 
Exemplified by AlphaGo \citep{alphago-2017silver}, self-play learning frameworks \citep{self-play-fu2023, self-play-chen2024} are proposed to improve LLMs' performance. 
Compared to classic methods \citep{bilateral-2024-wang}, LLMs based methods can perform inference across a wider variety of scenarios \citep{swiftsage-yao2023}, even with some ability of theory of mind \citep{agent-tom-2023-zhou} to infer other agent’s mental states. However, the inherit biases that can potentially limit their inferential abilities \citep{storytelling-2023-xie, opinion-dynamics-2024-chuang, finetune-prompt-wang2024}.
Works have also explored utilising LLMs as the environment \citep{zhang2024languageguided} or update actions \citep{action-learning-2024-zhao} for agents.


\paragraph{Optimisation for Complicated Tasks}
Alignment through human feedback offers more consistent training compared to reinforcement learning \citep{learning-to-learn-2024liang}, but obtaining this feedback can be expensive. Therefore, approaches like self-instruct \citep{self-instruct-wang2022, liu2023agentbench}, self-reflect \citep{tree-of-thoughts-yao2023}, self-alignment \citep{self-alignment-sun2023, rain-li-2024}, and few-shot planning \citep{song2023llmplanner}, have been introduced.
These approaches was also adapted to search for optimal tools \citep{anytool-du2024}, and interact with grounded environments \citep{autoplan-2023-ouyang, narrate-2024-ismail}. 
We were also inspired by stochastic search methods for 
robots in planning optimal strategies in complex environments \citep{DBLP:journals/ijrr/GammellBS20, learning-to-learn-2024liang}, shares many similarities with optimisation tasks for agents \citep{progprompt-2023-singh}.

\section{Conclusion}

This study introduces \agentname, a framework designed to address the challenges faced by LLM-driven agents in Murder Mystery Games (MMGs) through sensor-based state modelling, information gain-guided questioning, and goal-oriented suspect pruning. Supported by the WellPlay dataset for systematic evaluation, \agentname\ demonstrates improvements in strategic reasoning, interaction efficiency, and human-like behavioural quality. These results highlight the potential of integrating structured search and adaptive interaction techniques to advance AI agents in complex, dynamic multi-agent environments.

\bibliography{tacl2021}
\bibliographystyle{acl_natbib}

\newpage
\appendix

\begin{table*}[h!]
\centering
\renewcommand{\arraystretch}{1.1} 
\setlength{\tabcolsep}{5pt} 
\resizebox{0.9\linewidth}{!}{%
\begin{tabular}{|c|c|p{13cm}|}
\hline
\textbf{Type} & \textbf{Aspects} & \textbf{Examples (The correct answer has been highlighted in bold.)} \\ 
\hline
\multirow{3}{*}{\begin{tabular}{c}\textbf{A}\\(score 10)\end{tabular}} & \multirow{3}{*}{Who} & \textbf{Who killed Hans Li Morette?} \newline 
\textbf{A. Gale Li Morette} \quad B. Nurse head [Sylvia Costa] \quad C. Drake Li Morette \quad D. Frank Bijeli \\ 
\hline

\multirow{13}{*}{\begin{tabular}{c}\textbf{B}\\(score 5)\end{tabular}} & \multirow{2}{*}{How} & \textbf{How did Hans Li Morette die?} \newline 
\textbf{A. Shot to death} \quad B. Beaten to death \quad C. Poisoned to death \quad D. Drowned \\ 
\cline{2-3} 

& \multirow{2}{*}{Why} & \textbf{What was the motive behind the killing?} \newline 
A. Love killing \quad B. Vendetta \quad \textbf{C. Interest} \quad D. Accidental killing \\ 
\cline{2-3} 

& \multirow{2}{*}{Relationship} & \textbf{What is the relationship between the murderer and Hans Li Morette?} \newline 
A. Enemies \quad B. Colleague \quad C. Friend \quad \textbf{D. Wife} \\ 
\cline{2-3} 

& \multirow{2}{*}{Where} & \textbf{Where was Hans Li Morette killed?} \newline 
A. Emergency room \quad B. Johnson’s House \quad \textbf{C. Laboratory} \quad D. Dressing room \\ 
\cline{2-3} 

& \multirow{2}{*}{When} & \textbf{When was Hans Li Morette killed?} \newline 
A. This afternoon from 5:00 to 5:30 \quad \textbf{B. This afternoon from 6:30 to 7:00} \quad C. Tonight from 7:00 to 7:30 \quad D. This morning from 6:30 to 7:00 \\ 
\cline{2-3} 

& \multirow{2}{*}{Suspect} & \textbf{Please select the two most suspicious people:} \newline 
\textbf{A. Gale Li Morette} \quad B. Sylvia Costa \quad C. Drake Li Morette \quad D. Frank Bijeli \\ 
\hline

\multirow{11}{*}{\begin{tabular}{c}\textbf{C}\\(score 2)\end{tabular}} & \multirow{3}{*}{Three relationships} & \textbf{What is the non-existent relationship between Hans Li Morette and Andrew Paloski?} \newline 
A. Colleague \quad B. Mentor \quad C. Jealous \quad \textbf{D. Future daughter-in-law} \\ 
\cline{2-3} 

& \multirow{4}{*}{Two relationships} & \textbf{What is the relationship between Father Tom and Tony?} \newline 
\textbf{A. Tony is manipulated by x and deceived by x of Father Tom} \quad B. Father Tom is an authority over x and student of Tony \quad C. Father Tom is a student and ex-girlfriend of Tony \quad D. Father Tom is an ex-girlfriend and admired by x of Tony \\ 
\cline{2-3} 

& \multirow{4}{*}{One relationship} & \textbf{What is the relationship between Father Tom and Drake Li Morette?} \newline 
\textbf{A. Drake Li Morette is Father Tom's doctor} \quad B. Father Tom is helped by Drake Li Morette \quad C. Father Tom is the step-brother of Drake Li Morette \quad D. Father Tom hates Drake Li Morette \\ 
\hline
\end{tabular}%
}
\caption{Examples of Each Type of Our Evaluation Questions. For various types of questions, we assign different weights based on the original scoring system of the script. Specifically, Type A questions are valued at 10 points, Type B questions at 5 points, and Type C questions at 2 points.}
\label{tab:question_example}
\end{table*}

\section{The WellPlay Dataset} \label{WellPlay}

We recruited four human experts to read all the scripts and write the questions and answers. We conducted training sessions for them and organised game sessions to let them play and become familiar with the game flow. They were compensated at an hourly rate of \$31.92, with each narrative estimated to take about 5-20 hours to complete, depending on the complexity of the game.


\section{Human Subject Evaluation Setup} \label{human-setup}
Since the same player would know the killer's identity if they tried different models, we recruited four players to act as the murderer, each playing through three scripts (Sin, Deadly Fountain, Riverside Inn) across four methods, totalling 49 hours of gameplay. Each human player assumed the role of one character in the game, while the other characters were controlled by \agentname \ -powered agents. 

\begin{table}[h]
\centering
\resizebox{\linewidth}{!}{
\begin{tabular}{ccccc}
\toprule
\textbf{Evaluation} & \textbf{Werewolf} & \textbf{O-CoT} & \textbf{ThinkThrice} & \textbf{\agentname \ } \\
\midrule
Story Advancement  & $1.50_{\pm0.37}$ & $2.08_{\pm0.14}$ & $2.67_{\pm0.00}$ & $3.50_{\pm0.17}$ \\
Question Quality & $1.67_{\pm0.41}$ & $1.58_{\pm0.36}$ & $2.08_{\pm0.28}$ & $3.25_{\pm0.28}$ \\
Response Quality & $2.83_{\pm0.17}$ & $2.92_{\pm0.36}$ & $3.00_{\pm0.33}$ & $3.50_{\pm0.29}$ \\
Response Speed & $1.17_{\pm0.17}$ & $1.33_{\pm0.41}$ & $1.17_{\pm0.17}$ & $2.50_{\pm0.17}$ \\
Role Immersion & $2.25_{\pm0.36}$ & $2.17_{\pm0.5}$ & $2.33_{\pm0.24}$ & $2.83_{\pm0.29}$ \\
overall & $1.88_{\pm0.59}$ & $2.01_{\pm0.55}$ & $2.25_{\pm0.62}$ & $3.11_{\pm0.39}$ \\

\bottomrule
\end{tabular}}
\caption{Results of Agent-vs-Human Evaluation using Human-Centric metrics. }
\label{tab:human-survey}
\end{table}

To ensure fairness, the human players were given the same information and rules as the agents, maintaining consistent conditions across both evaluation scenarios. Both the human players and the agents shared the same objectives: to achieve victory in the game by fulfilling their individual objectives. This setup allowed for a direct comparison between human-agent and agent-agent interactions, providing insights into \agentname \ 's ability to adapt and perform in mixed environments.

Human players are given a questionnaire to assess their experience when playing with agents across five key dimensions: \emph{Story Advancement, Question Quality, Response Quality, Role Immersion,} and \emph{Response Speed}. Each dimension is rated on a scale from 1 to 5, with higher scores indicating stronger performance:

\subsection{Story Advancement}
\begin{tcolorbox}[title=Story Advancement Evaluation,
colback=white,
colframe=blue!75!black,
colbacktitle=blue,
coltitle=white,
breakable,
fonttitle=\bfseries]
\textbf{Score 5:} Actively gathers information and conducts in-depth analyses. Reasoning is rigorous, offering key insights that significantly advance the narrative.\\
\textbf{Score 4:} Proactively seeks information and performs reasonable analysis. Reasoning is generally accurate and supports narrative progression.\\
\textbf{Score 3:} Participates in information gathering but lacks analytical depth. Reasoning may be somewhat biased, offering moderate narrative progression.\\
\textbf{Score 2:} Shows limited initiative in gathering information, with superficial analysis. Reasoning is frequently flawed, hindering effective plot advancement.\\
\textbf{Score 1:} Does not gather information or attempt analysis. Unable to perform meaningful reasoning, leaving the narrative stagnant.
\end{tcolorbox}

\subsection{Question Quality}
\begin{tcolorbox}[title=Question Quality Evaluation,
colback=white,
colframe=green!75!black,
colbacktitle=green,
coltitle=white,
breakable,
fonttitle=\bfseries]
\textbf{Score 5:} Poses highly relevant, in-depth questions that directly propel the plot and uncover hidden clues.\\
\textbf{Score 4:} Most questions are relevant and yield useful information. Occasionally introduces innovative queries.\\
\textbf{Score 3:} Some questions are relevant but lack depth. Occasional repetition or irrelevant queries occur.\\
\textbf{Score 2:} Most questions bear little relevance, often fail to obtain valuable clues, and frequently repeat similar inquiries.\\
\textbf{Score 1:} Poses irrelevant or no questions, providing no support for narrative progression.
\end{tcolorbox}

\subsection{Response Quality}
\begin{tcolorbox}[title=Response Quality Evaluation,
colback=white,
colframe=red!75!black,
colbacktitle=red,
coltitle=white,
breakable,
fonttitle=\bfseries]
\textbf{Score 5:} Provides accurate and comprehensive responses fully aligned with the character’s background and narrative needs. Offers clear, valuable information.\\
\textbf{Score 4:} Generally provides accurate and sufficiently complete answers. Minor omissions occur but do not hinder overall understanding.\\
\textbf{Score 3:} Responses are basically accurate but sometimes incomplete or vague. May require further clarification by others.\\
\textbf{Score 2:} Responses are often inaccurate, incomplete, or misleading. Frequent clarification requests from others are necessary.\\
\textbf{Score 1:} Responses are unrelated or outright refusals. Offers no assistance in understanding the narrative.
\end{tcolorbox}

\subsection{Role Immersion}
\begin{tcolorbox}[title=Role Immersion Evaluation,
colback=white,
colframe=purple!75!black,
colbacktitle=purple,
coltitle=white,
breakable,
fonttitle=\bfseries]
\textbf{Score 5:} Fully embodies the character. Actions, dialogue, and emotional expressions are authentically aligned with the character’s traits, leaving a strong impression.\\
\textbf{Score 4:} Frequently demonstrates character traits. Occasionally shows minor deviations, but emotional expression is largely appropriate.\\
\textbf{Score 3:} Occasionally exhibits character traits, but inconsistently. Emotional expression is average, and immersion is limited.\\
\textbf{Score 2:} Characterization is unclear; actions and dialogue often stray significantly from the intended character profile. Emotional expression is awkward or absent.\\
\textbf{Score 1:} Demonstrates no character immersion; actions and dialogue contradict character traits. No emotional expression is evident.
\end{tcolorbox}

\subsection{Response Speed}
\begin{tcolorbox}[title=Response Speed Evaluation,
colback=white,
colframe=orange!75!black,
colbacktitle=orange,
coltitle=white,
breakable,
fonttitle=\bfseries]
\textbf{Score 5:} Responds promptly with no perceptible delay. The thought process is fluid and maintains a brisk, engaging pace.\\
\textbf{Score 4:} Responds relatively quickly with brief, occasional delays. The thought process is mostly smooth and well-paced.\\
\textbf{Score 3:} Moderate response speed with some noticeable pauses. The reasoning process is acceptable but not seamless.\\
\textbf{Score 2:} Responses are slow with frequent delays. The reasoning process is disjointed, disturbing the overall gameplay flow.\\
\textbf{Score 1:} Extremely slow responses or prolonged silence. The reasoning process halts, severely impeding the game’s progress.
\end{tcolorbox}

\section{Implementation}\label{appendix:Implementation}

\subsection{Implementation Details} \label{appendix:implement details}

We accessed GPT-3.5 and the GPT Embedding Model via the Azure API \footnote{https://learn.microsoft.com/en-us/azure/ai-services/openai/concepts/models}, using gpt-35-turbo-16k 0613 and text-embedding-ada-002. 
For retrieval enhancement, we utilised the FAISS\footnote{https://github.com/facebookresearch/faiss} library to construct a vector database, creating FAISS indices using the L2 distance metric. 
Scripts were divided into segments, with each segment having a maximum length of 50 tokens. For dialogue records, a question-and-answer pair was stored as a single segment. During retrieval, the maximum script and dialogue lengths included in the prompt were set to 4000 tokens. For evaluation, these maximum lengths were increased to 5000 tokens. Additionally, the hyperparameters were set with $\epsilon$ equal to 0.1 and $\beta$ equal to 0.2.

\begin{table*}[h!]
\centering
\resizebox{0.9\linewidth}{!}{
\begin{tabular}{ccccccccc}
\toprule
\multicolumn{1}{c}{\multirow{2}{*}{\textbf{Script}}} & \multirow{2}{*}{\textbf{\#Tokens}} & \multirow{2}{*}{\textbf{Stage}} & \multicolumn{1}{c}{\multirow{2}{*}{\textbf{PP}}} & \multicolumn{1}{c}{\multirow{2}{*}{\textbf{OP}}} & \multicolumn{4}{c}{\textbf{Agent's Response After Playing the Game}}    \\ \cmidrule(lr){6-9}
\multicolumn{1}{c}{}         &      & \multicolumn{1}{c}{}       & \multicolumn{1}{c}{}         & \multicolumn{1}{c}{}            & \multicolumn{1}{c}{\textbf{Werewolf}} & \multicolumn{1}{c}{\textbf{O-CoT}} & \multicolumn{1}{c}{\textbf{ThinkThrice}} & \multicolumn{1}{c}{\textbf{\agentname \ }} \\
\midrule
\multirow{2}{*}{\shortstack[c]{\emph{Death Wears White}\\ \emph{(9 players, 1 victim)}}} & \multirow{2}{*}{3,190}
&   Gameplay   & $-$ & $-$ & $3.797$ & $4.643$ & $3.992$ & $1.782 $ \\
& & Evaluation    & $0.349$ & $1.433$ & $0.815$ & $0.807$ & $0.790$ & $0.799$ \\
\midrule
\multirow{2}{*}{\shortstack[c]{\emph{Ghost Revenge}\\ \emph{(7 players, 3 victims)}}}  & \multirow{2}{*}{5,487}
&   Gameplay   & $-$ & $-$ & $6.702$ & $8.231$ & $7.056$ & $3.152 $ \\
& & Evaluation    & $0.418$ & $1.945$ & $1.006$ & $1.004$ & $0.995$ & $1.012$ \\
\midrule
\multirow{2}{*}{\shortstack[c]{\emph{Danshui Villa}\\ \emph{(7 players, 2 victims)}}}  & \multirow{2}{*}{5,111}
&   Gameplay   & $-$ & $-$ & $5.215$ & $6.262$ & $5.449$ & $2.440 $ \\
& & Evaluation    & $0.351$ & $1.415$ & $0.834$ & $0.816$ & $0.833$ & $0.825$ \\
\midrule
\multirow{2}{*}{\shortstack[c]{\emph{Unfinished Love }\\ \emph{(7 players, 2 victims)}}}  & \multirow{2}{*}{2,501}
&   Gameplay   & $-$ & $-$ & $3.945$ & $4.754$ & $4.237$ & $1.858 $ \\
& & Evaluation    & $0.230$ & $0.872$ & $0.487$ & $0.502$ & $0.499$ & $0.494$ \\
\midrule
\multirow{2}{*}{\shortstack[c]{\emph{Cruise Incident}\\ \emph{(5 players, 1 victim)}}}  & \multirow{2}{*}{1,262}
&   Gameplay   & $-$ & $-$ & $0.975$ & $1.146$ & $1.021$ & $0.448 $ \\
& & Evaluation    & $0.064$ & $0.327$ & $0.162$ & $0.165$ & $0.164$ & $0.163$ \\
\midrule
\multirow{2}{*}{\shortstack[c]{\emph{Sin}\\ \emph{(4 players, 1 victim)}}}  & \multirow{2}{*}{2,121}
&   Gameplay   & $-$ & $-$ & $0.680$ & $0.833$ & $0.730$ & $0.320 $ \\
& & Evaluation    & $0.061$ & $0.287$ & $0.141$ & $0.140$ & $0.141$ & $0.142$ \\
\midrule
\multirow{2}{*}{\shortstack[c]{\emph{Deadly Fountain }\\ \emph{(4 players, 1 victim)}}}  & \multirow{2}{*}{1,852}
&   Gameplay   & $-$ & $-$ & $0.671$ & $0.810$ & $0.724$ & $0.316 $ \\
& & Evaluation    & $0.045$ & $0.207$ & $0.107$ & $0.111$ & $0.111$ & $0.109$ \\
\midrule
\multirow{2}{*}{\shortstack[c]{\emph{Unbelievable Incident}\\ \emph{(5 players, 1 victim)}}}  & \multirow{2}{*}{3,182}
&   Gameplay   & $-$ & $-$ & $1.304$ & $1.567$ & $1.367$ & $0.610 $ \\
& & Evaluation    & $0.077$ & $0.291$ & $0.159$ & $0.163$ & $0.161$ & $0.162$ \\
\midrule
\multirow{2}{*}{\shortstack[c]{\emph{Desperate Sunshine}\\ \emph{(4 players, 1 victim)}}}  & \multirow{2}{*}{3,370}
&   Gameplay   & $-$ & $-$ & $0.803$ & $0.972$ & $0.847$ & $0.372 $ \\
& & Evaluation    & $0.104$ & $0.383$ & $0.222$ & $0.221$ & $0.224$ & $0.220$ \\
\midrule
\multirow{2}{*}{\shortstack[c]{\emph{Riverside Inn}\\ \emph{(4 players, 1 victim)}}}  & \multirow{2}{*}{1,909}
&   Gameplay   & $-$ & $-$ & $0.633$ & $0.762$ & $0.682$ & $0.297 $ \\
& & Evaluation    & $0.055$ & $0.223$ & $0.120$ & $0.124$ & $0.121$ & $0.123$ \\
\midrule
\multirow{2}{*}{\shortstack[c]{\emph{Solitary Boat Firefly}\\ \emph{(6 players, 4 victims)}}}  & \multirow{2}{*}{8,893}
&   Gameplay   & $-$ & $-$ & $7.799$ & $9.257$ & $8.252$ & $3.604 $ \\
& & Evaluation    & $0.380$ & $1.571$ & $0.816$ & $0.811$ & $0.814$ & $0.823$ \\
\midrule
\multirow{2}{*}{\shortstack[c]{\emph{Manna}\\ \emph{(6 players, 3 victims)}}}  & \multirow{2}{*}{9,028}
&   Gameplay   & $-$ & $-$ & $5.805$ & $6.925$ & $6.108$ & $2.665 $ \\
& & Evaluation    & $0.444$ & $1.864$ & $0.944$ & $0.965$ & $0.937$ & $0.953$ \\
\midrule
\multirow{2}{*}{\shortstack[c]{Overall}}  & \multirow{2}{*}{47,906}
&   Gameplay   & $-$ & $-$ & $38.329$ & $46.162$ & $40.464$ & $\bm{17.862}$ \\
& & Evaluation    & $2.578$ & $10.819$ & $5.813$ & $5.831$ & $5.789$ & $5.825$ \\
\bottomrule
\end{tabular}}
\caption{Compare the costs in US dollars(\$) of calling OpenAI API across multi-agent algorithms in MMGs setting, with Gameplay and Evaluation Stage. $\#Tokens$ represent the average length of each character's script. Costs are reported for one complete gameplay and one evaluation process for each script. }

\label{tab:cost-results}                       
\end{table*}

\paragraph{Experiment}
Following the results of our ablation studies, the gameplay phase was structured to ask one question per round over three rounds. After the game concluded, the evaluation phase consisted of three separate evaluations, with the final results being the average of these evaluations. 

\label{appendix:implementation}

\subsection{Overall Performance Computing} \label{appendix:Overall Performance Computing}
In calculating the overall score for performance, we have employed both the weighted mean and the weighted standard deviation. The weighted mean is computed by considering the count of questions for a specific category across various scripts as the weight. For the overall score, the total possible score for each script serves as the weight. This method allows us to adjust the influence of each category and script based on its significance and scale, thus providing a more nuanced and accurate reflection of performance.

The weighted mean is calculated as:
\[
\bar{x}_w = \frac{\sum_{i=1}^{n} (w_i \cdot x_i)}{\sum_{i=1}^{n} w_i}
\]

The weighted standard deviation, which measures the spread of the scores, is calculated using the weighted variance:
\[
s_w^2 = \frac{\sum_{i=1}^{n} w_i \cdot (x_i - \bar{x}_w)^2}{\sum_{i=1}^{n} w_i}
\]
And the weighted standard deviation is the square root of the weighted variance:
\[
s_w = \sqrt{s_w^2}
\]

\end{document}